\documentclass[twoside,11pt,preprint]{article}

\usepackage{blindtext}
\usepackage{float}

\usepackage[abbrvbib, preprint]{dmlr2e}
\usepackage{xcolor}
\usepackage{multirow}
\usepackage{booktabs}

\usepackage{lastpage}

\firstpageno{1}

\begin{document}

\title{Brevity is the soul of wit: Pruning long files for code generation}

\author{\name Aaditya K. Singh \email aaditya.singh.21@ucl.ac.uk \\
       Gatsby Unit, UCL
       \AND
       \name Yu Yang \email yuyang@cs.ucla.edu \\
       UCLA
       \AND
       \name Kushal Tirumala \email ktirumala@meta.com \\
       Meta FAIR
       \AND
       \name Mostafa Elhoushi \email m.elhoushi@ieee.org \\
       Meta FAIR
       \AND
       \name Ari S. Morcos \email ari@datologyai.com \\
       DatologyAI
}

\maketitle

\begin{abstract}
Data curation is commonly considered a ``secret-sauce'' for LLM training, with higher quality data usually leading to better LLM performance.  Given the scale of internet-scraped corpora, data pruning has become a larger and larger focus.
Specifically, many have shown that de-duplicating data, or sub-selecting higher quality data, can lead to efficiency or performance improvements. Generally, three types of methods are used to filter internet-scale corpora: embedding-based, heuristic-based, and classifier-based. In this work, we contrast the former two in the domain of finetuning LLMs for code generation. We find that embedding-based methods are often confounded by length, and that a simple heuristic---pruning \textit{long} files---outperforms other methods in compute-limited regimes. Our method can yield up to a 2x efficiency benefit in training (while matching performance) or a 3.5\% absolute performance improvement on HumanEval (while matching compute). However, we find that perplexity on held-out long files can increase, begging the question of whether optimizing data mixtures for common coding benchmarks (HumanEval, MBPP) actually best serves downstream use cases. Overall, we hope our work builds useful intuitions about code data (specifically, the low quality of extremely long code files) provides a compelling heuristic-based method for data pruning, and brings to light questions in how we evaluate code generation models.
\end{abstract}


\section{Introduction}

Large language models (LLMs) trained on internet-scale corpora showcase impressive capabilities \citep{gpt3}. With their increasing prevalence, one of the main paradigms has become training models on more and more data \citep{touvron2023llama1}. In fact, some go on to argue that the biggest factor in modern LLM training pipelines is the data, rather than architectural differences or other hyperparameters \citep{theit}. One of the key findings of modern LLMs has been that mixing in large amounts of diverse, but lower quality, data with higher quality data sources (e.g., Wikipedia) leads to improved performance \citep{touvron2023llama1}. Currently, the scale of internet-scale corpora \citep{together2023redpajamav2} still exceeds the training token budgets of most models, begging the question of how to sub-select data to use for training given a fixed compute budget.

In answer to this question, many recent works have focused on \textit{data pruning}. Data pruning aims to subselect data in order to 1. increase training efficiency (same number of epochs, but each epoch is smaller because it's on a pruned dataset) or 2. increase performance (compute-controlled, meaning more epochs). The former may have more impact in low-resource or academic settings, whereas the latter may have more impact on frontier LLMs. While historically models have been trained in the sub-1-epoch regime, recent work has shown that multiple epochs can be performed without sacrificing performance \citep{muennighoff2023scaling_epochs}, and that perhaps sub-selecting data and epoch'ing on it outperforms training less than one epoch on the full dataset \citep{tirumala2023d4}. These findings emphasize the need for good data pruning methodologies for training LLMs.

While the field of data pruning in machine learning has a long history with many types of methods \citep{wang2023data}, we focus here on data pruning approaches for large-scale language-like datasets,\footnote{The scale of these datasets can make many classical pruning techniques infeasible.} which loosely fit into three categories. \textit{Embedding}-based pruning instead uses a self-supervised pre-trained embedding network to embed all data points, and then a heuristic to prune examples based on their embeddings (e.g., points that are within $\epsilon$ of each other \citep{abbas2023semdedup}). \textit{Heuristic}-based pruning involving using human intuition about datasets to prune data points, such as removing files containing ``lorem ipsum'' \citep{c4} or removing nearly duplicated training data by comparing n-gram statistics \citep{lee2022deduplicating}. \textit{Classifier}-based pruning trains a classifier to identify ``good'' data from ``bad'' data, and then prune the bad data. Of course, classifier-based pruning requires access to some form of label, which can make it difficult to apply; in practice, some heuristic is used to identify ``good'' data (e.g., webpages referenced on Wikipedia \citep{touvron2023llama1}) and a random sample of the rest is used as ``bad'' data.\footnote{One could consider improving this scheme by some version of self-distillation \citep{pham2022revisiting_self_distillation}.} Each of these methods comes with its own pros and cons: \textit{embedding}-based methods rely on good embedding networks, \textit{heuristic}-based methods rely on good human intuition, and \textit{classifier}-based methods rely on labeled ``good'' data.

In this work, we illustrate a simple, yet strong, heuristic-based method in the domain of finetuning LLMs for code generation, which outperforms existing embedding-based methods. We choose code generation as our domain of study due to its increasing importance as an end use case for LLMs \citep{rozière2024code} and in line with recent works on embedding-based data pruning \citep{yang2023scip}. In this domain, we demonstrate that embeddings used in prior work are confounded by document length, which we use to motivate our length heuristic for data pruning. Surprisingly, we find that \textit{long} files, which are often believed to be ``higher quality'' \citep{longpre2023pretrainers}, are often actually very low quality (in the domain of code), yet make up disproportionately large amounts of training datasets due to their length. By pruning long files, we show improvements in compute-limited regimes on the popular coding benchmarks of HumanEval \citep{chen2021humaneval} and MBPP \citep{austin2021program_mbpp}. However, we find these benefits to be minimal in compute-rich regimes involving more than 2 epochs, with aggressive pruning methods overfitting to shorter files, thus emphasizing the recent emerging notion that data pruning methods may not be compute-agnostic \citep{goyal2024scaling}. We hope our findings motivate practitioners to consider simple heuristics more, especially document length as it directly impacts prevalence in training data, and that our work adds to the increasing discourse on data pruning methodologies.

\section{Related work}

\textbf{Embedding-based data pruning} was first shown to work at scale by \citet{sorscher2023neural}, who introduced SSL-prototypes and demonstrated it's efficacy on Imagenet-scale pretraining for vision. Subsequent work, SemDeDup \citep{abbas2023semdedup}, applies the embedding-based data pruning paradigm to language modeling, showing gains when training OPT models \citep{zhang2022opt} up to 1.3B parameters on the C4 dataset \citep{c4}. D4 \citep{tirumala2023d4} combined Semdedup with SSL-prototypes at even larger scales (up to 6.7B parameter OPT models), showing persistent benefits for language modeling. More recently, SCIP \citep{yang2023scip} used embedding-based data pruning in the fine-tuning stage for code generation language models, showing performance benefits on common benchmarks \citep{chen2021humaneval, austin2021program_mbpp}.

\noindent
\textbf{Heuristic-based data pruning} is perhaps the oldest approach in data curation for language models. Heuristics can range from simple string-based removal, such as removing documents containing boilerplate ``lorem ipsum'' or bad words\footnote{\url{https://github.com/LDNOOBW/List-of-Dirty-Naughty-Obscene-and-Otherwise-Bad-Words}} \citep{c4}, to more sophisticated recent techniques for removing near-duplicates using MinHash \citep{lee2022deduplicating}. Recent work \citep{sharma2024text} has shown that such heuristics can be combined to provide a useful notion of text quality to be used for pruning. One issue with heuristic-based methods is that they are often customized to the underlying data domain---for example, \citet{c4} remove all documents containing curly braces to purposefully \textit{remove} code data. Obviously, such a method would not extend to domains such as code generation. Notably, it's common for prior work \citep{c4} to remove \textit{short} documents (fewer than 3 sentences), but to our knowledge removing \textit{long} documents has not been studied. Here, we show that long documents in code data are often ``low quality'' and that removing them can lead to efficiency and performance gains.

\noindent
\textbf{LLMs for code generation} became popular with the advent of generally capable language models \citep{gpt3}  which could be finetuned for domain-specific use cases \citep{chen2021humaneval}. Recent work \citep{rozière2024code} showed that finetuning a pretrained Llama model can outperform, at a fixed compute budget, training from scratch on code data \citep{alphacode, fried2023incoder, li2023starcoder}, which motivates our approach. In this setting, LLMs may be finetuned on domain-specific code datasets, such as The Stack \citep{kocetkov2022stack}. Models are evaluated across many benchmarks, with HumanEval \citep{chen2021humaneval} and MBPP \citep{austin2021program_mbpp} being two of the most common. We refer readers to \citet{xu2022survey} and \citet{rozière2024code} for more extensive related work on LLMs for code generation.

\section{Motivating a length-based heuristic for code data}
\label{sec:motivation}

\begin{figure}
    \centering
    \includegraphics[width=0.9\textwidth]{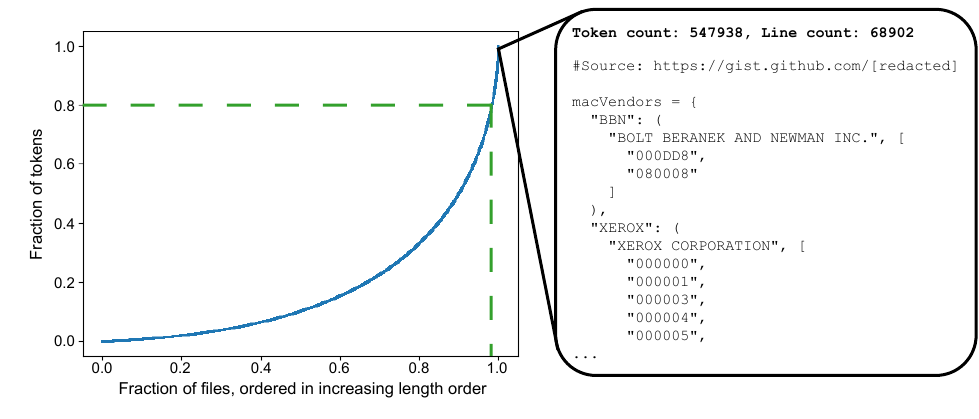}
    \vspace{-1.5em}
    \caption{Cumulative distribution of document lengths in the Python subset of the Stack dataset \citep{kocetkov2022stack}. Documents are ordered on the x-axis in increasing length order; y-axis shows fraction of whole dataset accounted for. Green lines highlight that the 2\% of longest files account for the last 20\% of tokens. Right shows an example long file (the second longest file). Long files tend to have very little useful signal, being full of large data arrays or ``spaghetti code''.}
    \label{fig:long_files_bad}
    \vspace{-2em}
\end{figure}

A common aphorism in software engineering is ``Don't Repeat Yourself'' (DRY), which is linked to the often frowned upon notion of ``spaghetti code''---long files with lots of repeated code. From a language modeling perspective, these tokens likely provide little-to-no signal for learning (similar to the decreased information when epoching \citep{muennighoff2023scaling_epochs}), and thus waste valuable compute. 

To get a sense of how such spaghetti code may be influencing our large-scale code datasets, we consider the distribution of document lengths in the Python subset of the popular Stack dataset \citep{kocetkov2022stack}. Replicating recent findings on other language datasets \citep{devries2023doc_lengths}, we find the distribution to be extremely right-skewed, with e.g., the longest 2\% of files making up a shocking 20\% of tokens. 

We qualitatively examine many of these files, with an example shown in Figure~\ref{fig:long_files_bad}. We find that most long Python files consist of large data arrays or very bad ``spaghetti code'':
The longest file is \textit{a single list} of names and phone numbers. Second longest is shown in Figure~\ref{fig:long_files_bad}b. Third longest is \textit{a single docstring} of chinese characters and anglicized pronunciations. Fourth longest is a \textit{game of hangman}, but coded without for-loops it would seem, resulting in essentially the same if-statement repeated thousands of times (for each index and word). Fifth longest is \textit{a single dictionary} of glyphs mapping to unicode characters. Each of these files has about 500,000 tokens.

Prior work has often pruned short files (e.g., shorter than 3 sentences as done by \citet{c4}), but we're unaware of existing work that specifically prunes \textit{long} files, which are often believed to be higher quality and useful for long-context training \citep{longpre2023pretrainers}. For the case of Python code, we see that this is false (at least for the longest files) motivating pruning long files as a heuristic.

\begin{figure}
    \centering
    \includegraphics[width=0.9\textwidth]{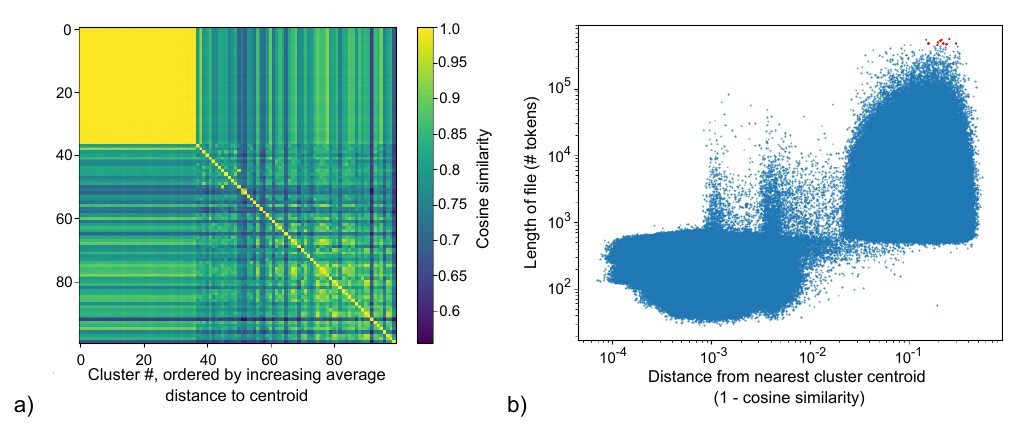}
    \vspace{-1.5em}
    \caption{Visualizations of StarEncoder embedding space. \textbf{a)} Cosine similarity between every pair of cluster centroids. Oddly, the first 37 cluster centroids are all \textit{very close} to each other. \textbf{b)} We visualize each document as an $(x,y)$ pair where the $x$-axis is the distance to nearest cluster centroid (computed as one minus the cosine distance), and the $y$-axis is the length of the file in tokens (using the Llama tokenizer). We immediately see the reason for the 37 super close clusters -- StarEncoder appears to tightly pack most documents below a certain length (bottom-left cloud of points), while longer documents tend to be more spread out in embedding space (top-right cloud of points). We also highlight the ten longest files in the dataset in red.}
    \vspace{-2em}
    \label{fig:starencoder_issues}
\end{figure}

Beyond this qualitative inspection of the data, we also find that existing state-of-the-art (SOTA) methods for data pruning in the domain of code generation \citep{yang2023scip} are actually confounded by file length. Specifically, the embedding space used for clustering by prior embedding-based methods is based off StarEncoder \citep{li2023starcoder}. Following the setup of \citet{yang2023scip}, we embedded all files in the Python subset of The Stack dataset using StarEncoder, and then performed K-means clustering with $K=100$ using the FAISS library \citep{johnson2019billion_faissgpu, douze2024faiss}. In Figure~\ref{fig:starencoder_issues}a, we find that many of the clusters are actually very close together, in terms of cosine similarity. In Figure~\ref{fig:starencoder_issues}b, we see a strong correlation between these clusters and file length.\footnote{We discuss further implications of this for embedding-based methods in Appendix~\ref{appx:tokens}.} Specifically, shorter files appear to be mapped to a very small and dense region of embedding space.\footnote{We also qualitatively inspected these files. Besides finding hundreds of thousands ($\approx3$\% of The Stack Python) of short, auto-generated Django migrations, we could not discern any correlates besides length.} As a result, well performing embedding-based methods, such as pruning points far from centroids \citep{yang2023scip}, often correlate to pruning long files, further motivating the use of this heuristic.

\section{Methods}
\label{sec:methods}

We finetuned Llama2 7B models on pruned versions of the Python subset of The Stack dataset (which has 20.4B tokens). We start by separating a small validation set, controlling for document length, to use in Section \ref{sec:case_study:eval}. To get a notion of variance in the results, we employed a boostrapping framework. Specifically, we conducted experiments on 3 random 50\% subsets of The Stack Python dataset. We applied each pruning technique to each subset independently, then trained models on each subset. We then compute standard error bars over the 3 subsets to get a sense of performance differences between ``different'' datasets. We hope this form of dataset bootstrapping becomes more common in the field, as we found it useful to get a sense of inherent noise.

As baselines, we consider no pruning, as well as the best performing method from \citet{yang2023scip}, which prunes 16\% of files from small clusters, and then 4\% of files that lie farthest from centroids (which we refer to as ``SCIP'').\footnote{See Appendix Table~\ref{tab:tokens_not_docs} for a reproduced table from \cite{yang2023scip} comparing various embedding-based pruning methods.} For our length pruning experiments, we consider pruning 10\%, 20\% or 50\% of \textit{tokens}, coming from the longest files.\footnote{As a sanity check, we also conducted some runs pruning from the shortest files, in smaller scale settings, and found that this was not effective.} Note the distinction between tokens and files; pruning 20\% of tokens from the longest files on the full dataset would correspond to pruning just 2\% of files (Figure~\ref{fig:long_files_bad}).

We train all models for 16,000 steps using a batch size of 1.3M tokens.\footnote{Hyperparameters, compute requirements, etc. are provided in Appendix~\ref{appx:hyperparams}.} This would correspond to just over 1 epoch on the full Stack dataset, which means it corresponds to just over 2 epochs for our baseline (recall: our bootstrapping uses random 50\% subsets of the full dataset), and just over 4 epochs for our strictest pruning condition (50\% of tokens). We evaluate intermediate checkpoints every 2,000 steps on the common code generation benchmarks HumanEval \citep{chen2021humaneval} and MBPP \citep{austin2021program_mbpp}.

\section{Results}
\label{sec:results}

In Figure \ref{fig:main_result}, we provide three views on our results:\footnote{Full training curves are provided in Appendix~\ref{appx:extra_results}, for maximum transparency.} for considering \textit{training efficiency}, we look at performance after a single epoch on all conditions (left column). We see that the 50\% length pruning run performs as well as no pruning, despite only seeing half the data, indicating a 2x training efficiency improvement. In the middle column (Figure \ref{fig:main_result}), we provide a comparison targeted at looking for \textit{performance improvements}. Namely, we consider the checkpoints after 8k steps in each condition (roughly 2 epochs for the 50\% length pruning condition, and roughly 1 epoch for the no pruning baseline). In this plot, we can see strong improvements (absolute: 3.5\%) from length pruning on HumanEval, and modest improvements on MBPP (absolute: 1.5\%).\footnote{Note, 8k steps corresponds to two epochs over the aggressively pruned dataset, which is a bit contrary to classical ``sub-1-epoch'' training recommendations. Our work is thus closer to more recent findings \citep{muennighoff2023scaling_epochs} which show multiple epochs can be beneficial for language models.} When training for longer (Figure~\ref{fig:main_result}, third column), however, we see that the delta between length pruning and baselines does not persist, replicating the findings of prior work showing that data pruning cannot be compute agnostic \citep{goyal2024scaling}. 

\begin{figure}[h]
    \centering
    \includegraphics[width=\textwidth]{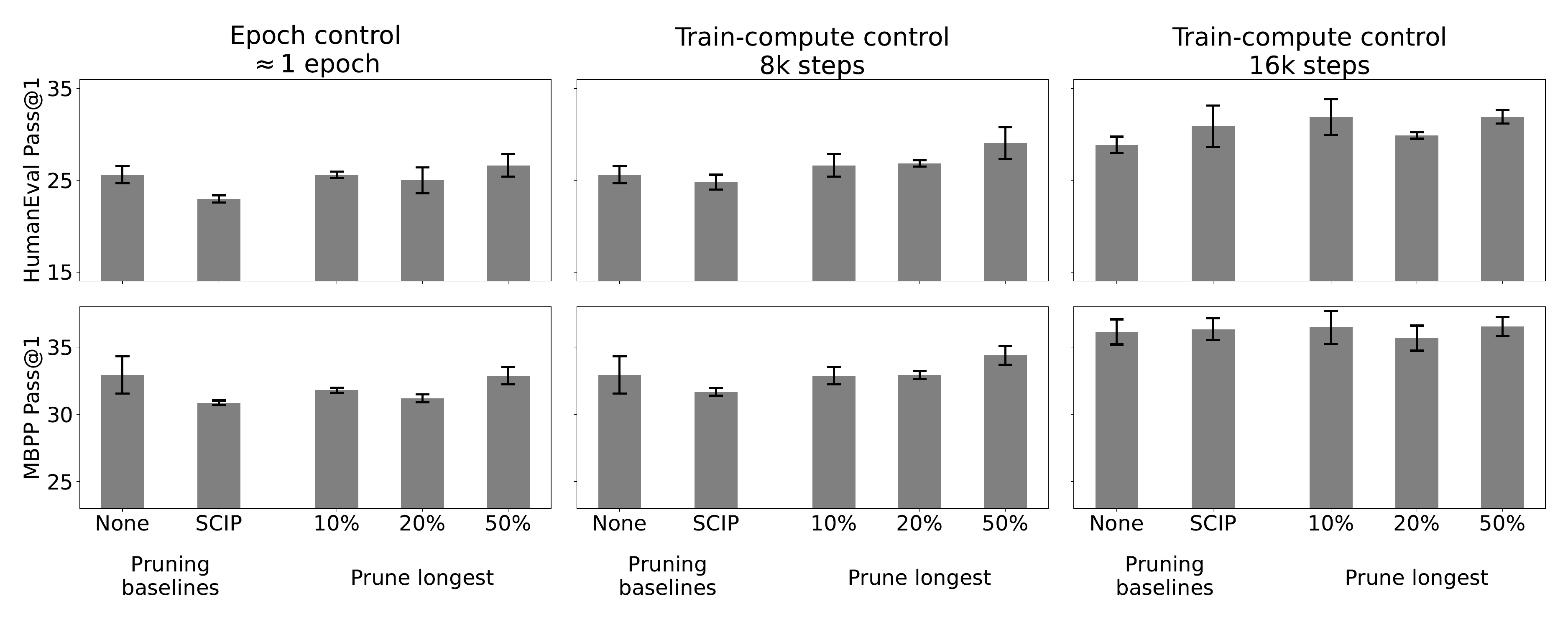}
    \vspace{-2.5em}
    \caption{Results across baselines and length pruning. All plots show the same 5 conditions: no pruning (``none'') and the best performing (embedding-based) method from \citet{yang2023scip} (``SCIP'') as baselines, and then pruning $P$\% of tokens from the longest documents (our method) for $P=10, 20, 50$. Leftmost column shows performance after roughly 1 epoch for each condition, which can be used to look at training efficiency. We note that aggressive length pruning ($P=50$\%) matches the performance of no pruning, indicating a 2x efficiency improvement. The latter two columns show compute-controlled experiments, where all conditions are trained on the same number of tokens (but possibly more epochs, e.g., the 50\% length pruning run will have seen twice as many epochs as no pruning). At 8000 steps (middle column), aggressive length pruning still seems to outperform no pruning and SCIP baselines, but this advantage diminishes at the larger compute scale (16000 steps, rightmost column). Error bars indicate standard error.}
    \vspace{-2em}
    \label{fig:main_result}
\end{figure}

Notably, length pruning outperforms SOTA embedding-based pruning methods (SCIP) in this domain. SCIP does not perform well at smaller compute scales (single-epoch regime, or even 8k steps). At larger compute scales (rightmost column) it surpasses no pruning on HumanEval, and matches on MBPP. In the SCIP paper \citep{yang2023scip}, models were trained for 60B tokens---it may be that SCIP performs better in these larger compute regimes. That said, we note that even in the larger compute regime, the length pruning method performs as well as SCIP.

\subsection{The importance of diverse evaluation}
\label{sec:case_study:eval}

So far, we've considered standard downstream benchmarks, where we found pruning long files leads to training efficiency and performance improvements. However, we note that most code generation benchmarks tend to be \textit{short-form} generation. When used as the sole signals of performance, these benchmarks could give us an overestimated benefit of pruning long files. To account for this, we look at perplexity on held out code, \textit{binned by length}. 

In Appendix Figure~\ref{fig:val_ppl}, we find that models can overfit to shorter code (indicated by increasing validation perplexity), especially for the most aggressive pruning strategies (where 50\% of tokens are removed, coming from the longest files) in the largest compute regimes (and thus the most epochs\footnote{Mirroring the findings of \citet{sorscher2023neural}, pruning long files could possibly perform \textit{better} if the source corpus is larger, as more steps could be taken on the pruned dataset without epoch'ing.}). These results offer a nice counterpoint to the work of \citet{rozière2024code}, who found that finetuning for long-context code generation often deteriorates performance on short-form code generation (as measured by HumanEval and MBPP). Both of these results taken together may point to an inherent tradeoff between short- and long-form code generation capabilities.\footnote{Related, we note that the performance improvements seen in \cite{li2023textbooks} may also be partially explained by the introduction of short-form synthetic code data.}

Depending on the downstream use case, such ``overfitting'' to short-form code generation may be okay---for example, if end users can only access a model up to its context length (4096 tokens, in our case), presumably practitioners will only care about generalization on code sequences up to the context length.

\section{Discussion}
\label{sec:discussion}

In this work, we showed that a simple heuristic-based data pruning method outperforms SOTA embedding-based methods when finetuning LLMs for code generation. Specifically, we found that file lengths are right skewed, leading to a small percentage of (typically ``low quality'') files making up a large percentage of code data. Removing these files from finetuning data yielded efficiency and performance improvements on downstream benchmarks in the compute-limited regime. We emphasize that these results were intended to introduce a new heuristic that we believe holds merit, and to contrast with embedding-based pruning methods; we are not asserting that length-based pruning is the ``best'' pruning method for general datasets and models. Overall, we hope our findings encourage practitioners to more carefully inspect subsets of data that have disproportionately large presence (i.e., long files), and also think about more principled ways of iterating on data pruning methodologies (e.g., bootstrapping for error bars, using a diverse suite of evaluations).

\subsection{Limitations}
\label{sec:limitations}

While our experiments show compelling performance improvements in the finetuning setting, due to compute constraints, we were not able to perform data ablations for pretraining models. It could be that longer files (e.g., with a lot of repeated code) could be useful at earlier points of training in incentivizing in-context learning abilities. Beyond this pre-/fine-tuning distinction, we also did not consider the use of classifier-based pruning. Quality classifiers are typically only used on CommonCrawl text data \citep{touvron2023llama2}. However, extending their usage to code data may end up filtering out many of the ``useless'' long files, such as those containing serial numbers (Figure~\ref{fig:long_files_bad}), thus reducing the need and/or impact of an explicit length-based heuristic filter. Finally, we note that our results are limited to the domain of code generation models. Further work would be needed to see if long files are generally lower quality data---in fact, we suspect this is generally not the case (e.g., for natural language corpora), but may be a factor in web scrapes for other subdomains.

\subsection{Finding ``high quality'' long-context code data}
\label{sec:long_context}

As LLMs are deployed more broadly, many model developers have focused on increasing the context length \citep{geminiteam2024gemini15}. When increasing context length, to make sure the model effectively learns to ``use it'', we need corresponding datasets with long documents. However, our findings on the low quality of longer Python files may beg the question of where such data could be found for the domain of code generation. To this end, we suggest the use of techniques such as In-context pretrain \citep{shi2023incontextpacking}, whereby similar documents are concatenated within a single context, to boost long-context capabilities. Given that much of The Stack is sourced from Github, model practitioners could also consider concatenating files from a single repository into a training sequence---for codebases utilizing object-oriented programming, this could be an especially compelling technique as some of the information needed to predict the next token in a given file may come from a different file.

\impact{Our work focuses on a simple new heuristic-based method for data pruning. We do not foresee negative societal consequences from this work, as our work does not unlock new capabilities. We believe the work has positive consequences for the research community, such as possibly enabling academic researchers to more efficiently train models and  urging data pruning researchers to consider simple methods more commonly in their research.}

\acks{The authors would like to thank William Held, Newsha Ardalani, Sho Yaida, Niladri Chatterji, Melanie Kambadur, Mike Rabbat, DJ Strouse, and Andrew Saxe for insightful discussions throughout the course of the work and feedback on the draft. All work was completed during A.K.S. and Y.Y.'s internships at Meta FAIR. The authors also thank Jeffery Yu for crucial latex help on earlier versions of the draft.}

\vskip 0.2in
\bibliography{references}

\clearpage

\appendix

\section{Embedding-based pruning methods on documents may prune vastly different numbers of tokens}
\label{appx:tokens}

\begin{table}
    \caption{(Reproduced with minor edits and permission from \citet{yang2023scip}) Pass@1 performance on HumanEval and MBPP for different pruning methods with 20\% \textit{documents} pruned. Each number here is the result of finetuning a 1.5B Llama1 \citep{touvron2023llama1} model on 67B tokens (multiple epochs). We refer readers to the original work for more detail. We add a \textbf{third row} quantifying the number of \textit{tokens} pruned by each method. Different methods often prune vastly different amounts of tokens.}
    \centering
    \resizebox{\textwidth}{!}{
    \begin{tabular}{lcccccccc}
    \toprule
      &  No & Random & SSL- & \multirow{2}{*}{SemDeDup} & \multirow{2}{*}{D4} & Small & Far from & Combined\\
      &  pruning & Pruning & Prototypes &  &  & Clusters & Centroids & Small+Far \\
    \midrule
    HumanEval & {25.0\%} & 24.0\% & 23.8\% & 20.7\% & 23.2\% & 23.2\% & 26.8\% & \underline{28.0\%} \\
    MBPP & 33.4\% & 31.9\% & 32.2\% & 32.4\% & 31.2\% & \underline{35.0\%} & 30.8\% & 33.0\% \\
    \midrule
    \textbf{Tokens pruned} & 0.0\% & 20.0\% & \textbf{4.1\%} & \textbf{5.7\%} & \textbf{5.8\%} & \textbf{25.1\%} & \textbf{20.2\%} & \textbf{24.6\%} \\
    \bottomrule 
    \end{tabular}
    }
    \vspace{-0.5em}
    \label{tab:tokens_not_docs}
\end{table}

In Table \ref{tab:tokens_not_docs}, we show the comparison of different embedding-based data pruning methods from \citet{yang2023scip}, where each method pruned 20\% of \textit{documents}. In the third row, we computed the percentage of \textit{tokens} pruned in each case. These results emphasize the difference between pruning in terms of documents vs. in terms of tokens for natural language datasets. In our work, we prune $P$\% of tokens, coming from the longest files.

\section{Hyperparameters}
\label{appx:hyperparams}

All models were finetuned using the AdamW \citep{adamw} optimizer ($\beta_1 = 0.9, \beta_2 = 0.95$) with weight decay of 0.1 and a learning rate of $1e-4$ that was cosine annealed down to $1e-5$. There was a linear warmup from 0 to $1e-4$ for the first 1000 steps, as is common practice. Model code was implemented using PyTorch \citep{pytorch} and fully-sharded data parallel \citep{zhao2023fsdp} was used to accelerate training. Each run took about 26 hours on 64 A100 GPUs.

\section{Additional results}
\label{appx:extra_results}

\begin{figure}
    \centering
    \includegraphics[width=\textwidth]{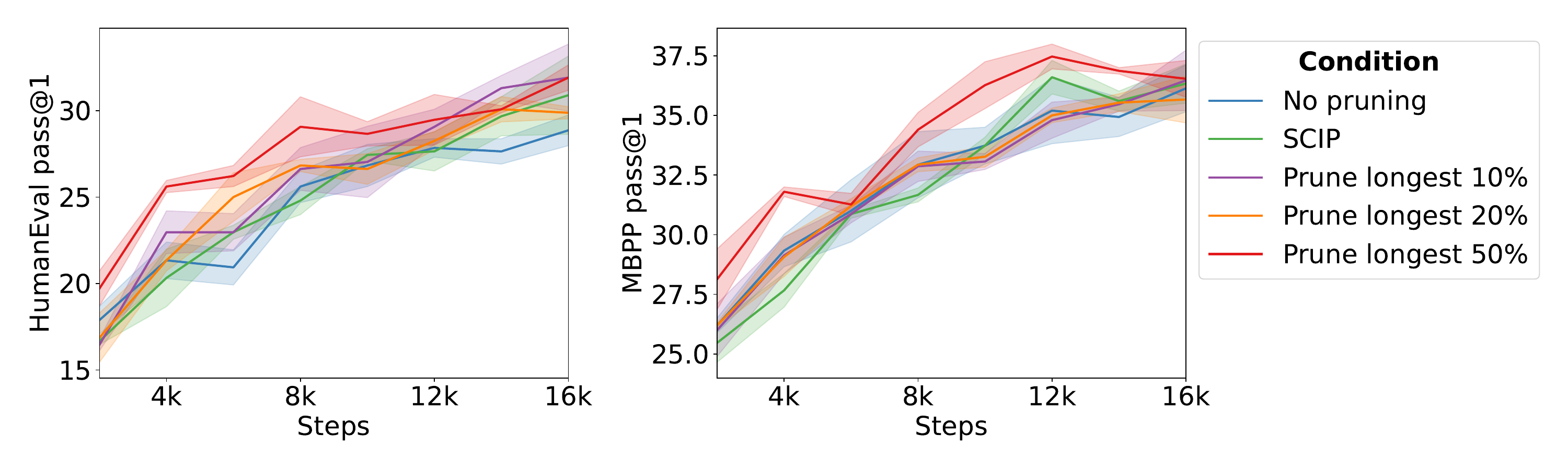}
    \vspace{-3em}
    \caption{Raw curves of evaluation benchmark performance over training. Shaded area indicates standard error. Note the results in Figure~\ref{fig:main_result} are taken from this plot, to more clearly show snapshots at various points during training.}
    \label{fig:raw_result}
\end{figure}

Full curves for eval performance (HumanEval and MBPP) throughout the finetuning process are provided in Figure~\ref{fig:raw_result}. 

\begin{figure}
    \centering
    a)\includegraphics[width=0.95\textwidth]{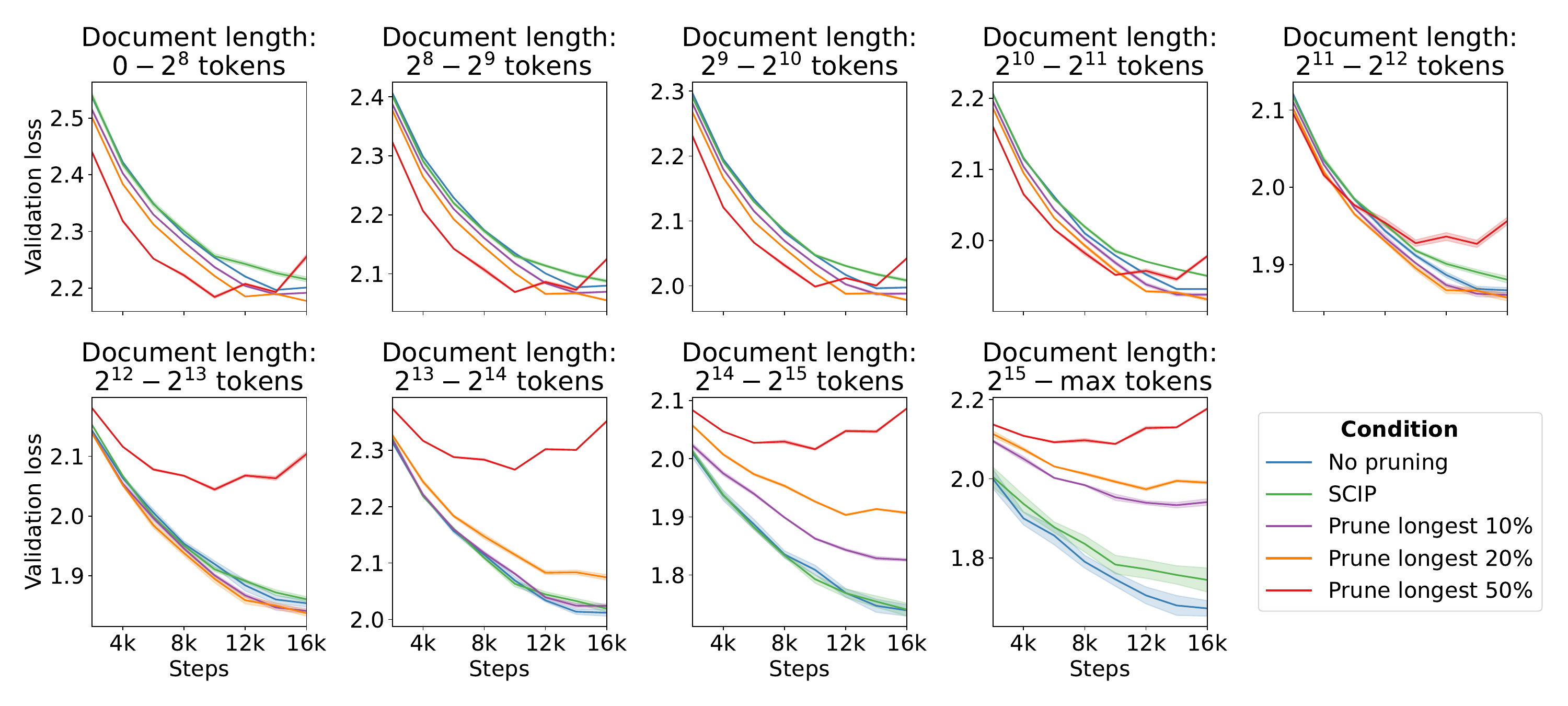} \\
    b)\includegraphics[width=0.95\textwidth]{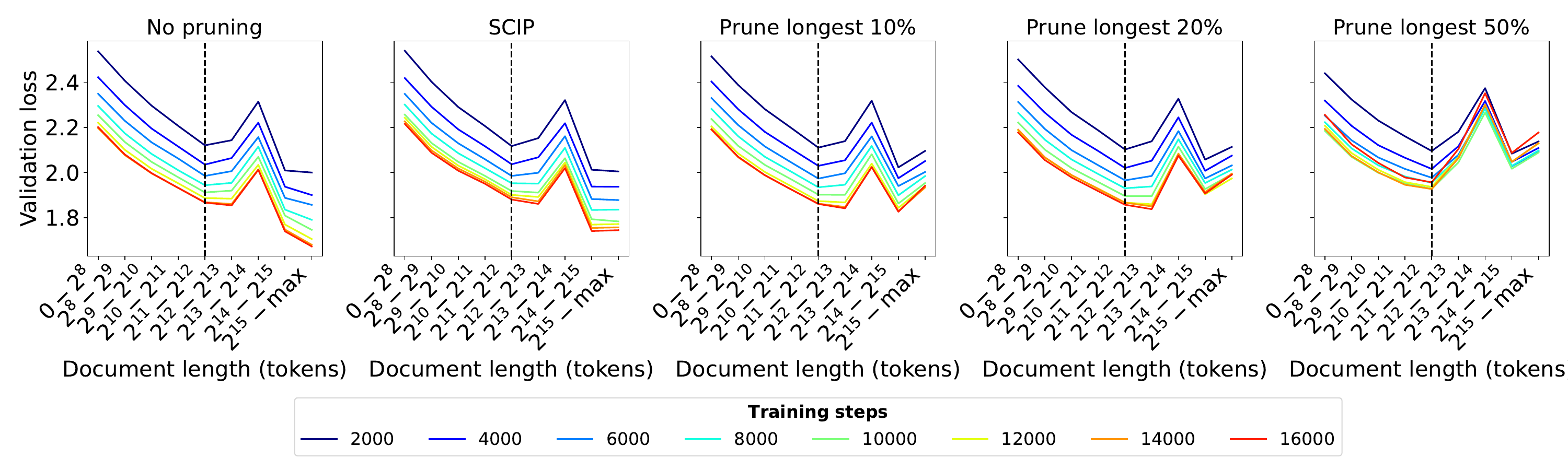}
    \caption{Two views on validation perplexity evolution on various length bins. \textbf{a)} We show the curves throughout training with each panel showing a different length bin. It's clear that the most aggressively pruned setting (red curve, Prune longest 50\%) overfits with multiple epochs (around 10k steps) on all lengths, and generally has worse perplexity on longer length documents---intuitively this makes sense since the longest file in the 50\% pruned subset was only $\approx3600$ tokens. Mirroring our results from Figure~\ref{fig:main_result}, this most aggressive pruning setting performs best in the compute-limited regime (up to 8k steps) on shorter documents (more akin to the downstream benchmarks HumanEval and MBPP). \textbf{b)} We instead focus on the perplexities across length bins, with each color showing a different step. Considering any time step of the ``no pruning'' baseline, we see that perplexities decrease with longer files, up to the context length (black dotted line), as we'd expect since there's more ``useful'' context enabling the model to lower perplexity. Beyond the context length, perplexities increase as expected (since all documents are being cut somewhere in the middle). Surprisingly, perplexities again decrease on the \textit{longest documents}, seeming to indicate that being split isn't affecting these documents. This is in line with our qualitative observations of repetitiveness in Section~\ref{sec:motivation}, and further motivates pruning these files.}
    \label{fig:val_ppl}
\end{figure}

Validation perlexities are calculated in the standard way by packing documents together (as done during training), breaking up into context length chunks (in our case, $2^{12}=4096$), and then calculating average per-token perplexity. We present two views on the results in Figure~\ref{fig:val_ppl}.

\end{document}